\documentclass{easychair}

\usepackage{doc}
\usepackage{float}
\title{Confidence-Credibility Aware Weighted Ensembles
of Small LLMs Outperform Large LLMs in
Emotion Detection}

\titlerunning{sLLMs Ensemble Outperform LLMs in Emotion Detection}
\authorrunning{Elgabry and Hamdi}

\begin{document}
\author{
Menna Elgabry\inst{1} \and
Ali Hamdi\inst{1}
}

\authorrunning{Elgabry and Hamdi}

\institute{
Department of Computer Science\\
MSA University\\
Giza, Egypt\\
\email{mennatullah.yasser@msa.edu.eg, ahamdi@msa.edu.eg}
}

\maketitle
\begin{abstract}
  This paper introduces a confidence-weighted, credibility-aware
ensemble framework for text-based emotion
detection, inspired by Condorcet’s Jury Theorem (CJT).
Unlike conventional ensembles that often rely on homogeneous
architectures, our approach combines architecturally diverse
small transformer-based large language models (sLLMs)—BERT,
RoBERTa, DistilBERT, DeBERTa, and ELECTRA—each fully
fine-tuned for emotion classification. To preserve error diversity,
we minimize parameter convergence while taking advantage of the unique biases of each
model. A dual-weighted voting
mechanism integrates both global credibility (validation F1-
score) and local confidence (instance-level probability) to
dynamically weight model contributions. Experiments on the
DAIR-AI dataset demonstrate that our credibility-confidence
ensemble achieves a macro F1-score of 93.5\%, surpassing
state-of-the-art benchmarks and significantly outperforming
large-scale LLMs, including Falcon, Mistral, Qwen, and Phi,
even after task-specific Low-Rank Adaptation (LoRA). With
only 595M parameters in total, our small LLMs ensemble
proves more parameter-efficient and robust than models up to
7B parameters, establishing that carefully designed ensembles
of small, fine-tuned models can outperform much larger LLMs
in specialized natural language processing (NLP) tasks such as
emotion detection.
\end{abstract}

\section{Introduction}
\label{sect:introduction}

Building intelligent interactive systems that can understand and respond to human emotions is a remarkable goal in computer science. As processing systems advance in their use of NLP technologies, correctly identifying and responding to the emotional complexities of a given text is essential for effective mental health applications \cite{dechoudhury2013depression}, personalized recommendations \cite{adomavicius2005recommender}, and human-computer interactions \cite{song2024nlp_hci_review}. Understanding this task is very difficult. Emotions cannot be understood through a simple framework of predetermined binary choices. Instead, they can be communicated implicitly, and they exist in varying amounts and may be represented in different forms, including irony, cultural references, and figurative language \cite{mohammad2013nrc}. Such complexities are recognized in the DAIR-AI benchmark dataset \cite{saravia-etal-2018-carer} that struggle to achieve reliable classification due to the inherent imbalance of the classification targets.

In text-based emotion detection (TED), the current approach involves the direct fine-tuning of large language models (LLMs) that have been previously trained on specific tasks. Although several models from the BERT family have been developed and deployed with success, their effectiveness has come to a standstill, especially with poorly distributed data. This limitation arises because a single model, no matter how sophisticated its architecture is, will always be biased toward the majority \cite{he2009imbalanced}. Consequently, interest has shifted toward the use of ensemble models because of their ability to combine the outputs of several models to deliver a more integrated and accurate classifier. An example of ensemble methods is the Condorcet's Jury Theorem (CJT), which states that a more accurate decision can be derived from a collection of independent and competent models than from any single member of the collection \cite{chung2023ensemble}.

However, the use of CJT in TED may not be optimal when Several ensembles are made of homogeneous models(multiple copies of the same architecture) for example, BERT fine-tuned with different seed initializations. The result of this is that the models reinforce each other and often make correlated errors, violating the independence assumption crucial to CJT’s success \cite{kuncheva2003diversity}. Moreover, Traditional aggregation by majority voting assumes that all models contribute equally to every prediction, which is clearly not the case. As, this equal treatment misses two important signals: (1) the overall reliability of a model, expressed in its evaluation metrics (e.g., F1-score on validation data), and (2) the confidence level for a prediction on a given case. Not considering these factors is likely to produce suboptimal ensemble decisions in tasks that require subtle discrimination, such as emotion classification.

In this work, we introduce a novel ensemble framework to mitigate these limitations using a dual-weighted voting mechanism. We perform a fine-grained six-class emotion classification setup where, unlike standard majority voting, our approach employs a confidence and credibility weighted voting system, where each model’s contribution is based on its voting confidence on the instance level and it's voting credibility on the dataset level. With this design, the ensemble can make predictions that are both  certain and reliable. This leads to better predictions for a range of emotions. We illustrate the overall effectiveness of our ensemble, as it consistently outperforms LLMs and achieves results that are within 0.2\% of a benchmark from a Q1 journal \cite{BARCENARUIZ2025114070}. These findings conclusively show that the improvements observed are due to the enhanced reliability of the suggested weighted voting system
\section{Background}
\subsection{ Transformers Architectures for Text Classification}

The transformer architecture introduced by Vaswani et al. \cite{vaswani2017attention} revolutionized modern Natural Language Processing (NLP) due to its self-attention mechanism, which evaluates the contextual relationships among all words in a sequence, resulting in a comprehensive understanding of the text. Transformer models are generally classified according to their pretraining objectives. For example, encoder models such as BERT \cite{devlin2019bert} are designed for natural language understanding tasks, as they generate bidirectional contextual representations, while decoder models such as GPT focus on text generation through auto regressive language modeling. On the other hand, encoder–decoder hybrids are optimized for sequence-to-sequence tasks such as translation and summarization, as demonstrated by models like T5 and BART \cite{wolf2020transformers}.

In the context of Text-based Emotion Detection (TED)—a classification-focused problem—encoder-based transformers tend to yield superior results due to their ability to identify and interpret intricate semantic and contextual nuances \cite{imran2024emotion}.

This study employs a varied ensemble of transformer models, integrating encoder based models, and comparing against modern decoder based Large Language Models (LLMs).

Encoder-Based Models: We utilize five members of the BERT family, chosen because of their proven efficiency and diverse architectures:

BERT (Bidirectional Encoder Representations from Transformers) \cite{devlin2019bert}: The foundational model, selected for its robust understanding of the bidirectional context.

RoBERTa (A Robustly Optimized BERT Pretraining Approach) \cite{liu2019roberta}: An optimized variant that removes the next-sentence prediction objective and incorporates more dynamic masking, frequently resulting in stronger performance in the General Language Understanding Evaluation (GLUE) and comparable evaluations.

DistilBERT \cite{sanh2019distilbert}: A distilled variant of BERT that has a 40\% reduction in size and 60\% increase in speed while retaining 97\% of its language understanding capabilities. Its modified layer structure introduces a unique "reasoning style" to the ensemble.

DeBERTa (Decoding-enhanced BERT with Disentangled Attention) \cite{he2021deberta}: This new version improves BERT by incorporating a disentangled attention mechanism and an advanced masking decoder, resulting in improved efficacy in numerous natural language understanding (NLU) tasks.

ELECTRA (Efficiently Learning an Encoder that Classifies Token Replacements Accurately) \cite{clark2020electra}: This model uses a novel pre-training task called replaced token detection, which is more sample-efficient than BERT's masked language modeling.

Decoder-Based Models (LLMs): we also compare our results against several decoder only LLMs finetuned for classification:

Phi \cite{li2023phi}: A high performance open-source model with a focus on common sense reasoning and logical understanding.

Falcon \cite{penedo2023falcon}: A suite of state-of-the-art LLMs trained in large and diverse corpora, known for their strong general capabilities.

Qwen (Qianwen) \cite{bai2023qwen}: A powerful multilingual LLM series with strong performance in a wide range of tasks.

Mistral \cite{jiang2023mistral}: An efficient model that uses grouped-query attention and sliding window attention to achieve high performance with lower computational cost.

OpenLLaMA \cite{geng2023openllama}: An open-source reproduction of the LLaMA model, providing a strong base for instruction tuning.

\subsection{Ensemble Learning and Condorcet's Jury Theorem}

Ensemble learning is a machine learning paradigm in which multiple models, or base learners, are combined to produce more robust and precise predictions \cite{zhou2012ensemble}. Its core hypothesis is that collective decisions can correct individual model errors, improving generalization through bias and variance reduction and better approximation of the underlying hypothesis space \cite{gupta2022ensembling}. Common ensemble techniques include bagging (e.g., Random Forests \cite{breiman2001random}), which reduces variance via bootstrapped sampling; boosting (e.g., AdaBoost \cite{freund1997decision}), which iteratively focuses on misclassified instances to reduce bias; and stacking \cite{wolpert1992stacked}, which employs a meta-learner to combine model outputs. In deep learning, neural network ensembles further enhance reliability and uncertainty estimation by averaging various solutions in the function space \cite{Thuy_2024}.

The theoretical foundation for the “wisdom of crowds” effect in ensembles is rooted in Condorcet’s Jury Theorem (CJT) \cite{condorcet1785essai, grofman1983thirteen}, which asserts that if each voter has a probability $(p>0.5)$ of being correct, the precision of the collective decision increases toward 1 as the number of voters grows. Similarly, ensemble accuracy improves when two conditions hold: (1) Competence—each model performs better than random chance $(p>0.5)$, ensuring its contribution adds signal rather than noise; and (2) Independence—model errors are uncorrelated, allowing one model’s correct predictions to offset mistakes from another.

A major limitation in neural ensemble design arises when this independence assumption is violated. Ensembles built from identical architectures (e.g., multiple BERT variants) finetuned on the same data often converge to similar solutions, producing highly correlated predictions and reduced ensemble gains \cite{rame2021dice}. Thus, ensuring diversity among base learners—through variations in architecture, data subsets, or training objectives—is essential for maximizing ensemble effectiveness \cite{wood2023unified}.

\section{Related Work}

In recent years, emotion detection has become a target of considerable attention,with advancements made across a multitude of different types, including text, speech, and multimodal approaches.

1. Bárcena Ruiz and Gil Herrera \cite{BARCENARUIZ2025114070}  have adapted an approach based on Condorcet's Jury Theorem (CJT) for Emotion Detection in Texts in Spanish and English. The primary approach is assembling a "jury" of complementary BERT-based transformers (BERT, RoBERTa, and DistilBERT), each independently fine-tuned on emotion datasets. The justification is that a majority vote from a set of competent and independent models will likely give better accuracy, even in instances when models are weak due to incomplete training (which for the purpose of this analogy meant training without certain emotions). Their major algorithmic contribution is the design of three ensemble aggregation algorithms. Jury Classic (JC), which uses simple majority voting, Jury Adaptive (JA), which uses a form of model reliability scoring, and Jury Dynamic (JD), a more sophisticated ensemble which applies "Jury Memory" to model reliability scoring. The algorithms aim to achieve a majority vote on each text input. In the absence of a majority, they are designed to make a random choice from the models (if memory is empty) or the highest ranked model based on historical votes. They tested this in Spanish and English on a number of datasets (DAIR-AI/emotion, SemEval 2018, XED).
The ensembles constructed using CJT showed a notable advantage over the different models, especially when the amount of training data was restricted. The ensembles considerably reduced the loss of efficiency that single models experienced, attaining higher F1-scores in the process. The study also showed the strong performance of RoBERTa in the ensemble and highlighted the challenge of data scarcity, especially for emotion detection.

2. Hamad et al. \cite{hamad2024asemenhancingempathychatbot} proposed ASEM (Attention-based Sentiment and Emotion Modeling) model, for enhancing empathy in open-domain chatbots. The foundation of their work derives from applications of sentiment analysis (positive, negative, and neutral) and how accurately sentiment analysis measures empathy in understanding the user’s emotional state. For example, emotions like joy and surprise are different, but can be categorized as the same sentiment. To improve analysis, ASEM uses a hierarchical model where analysis of a given text in the conversation begins with sentiment, and subsequently performs emotions analysis. ASEM uses a "Mixture of Experts" architecture with hierarchically integrated multiple encoders which provide different sentiment-specific feature representations. These representations are then passed to the specialized attention module which determines the sentiment of a given text and assumes the role of a context synthesizer by creating a representation that is enriched in the semantic and emotional space. The model predicts the emotion of the text and then generates a response using the corresponding decoder. ASEM is evaluated on Empathetic Dialogues (ED) and DailyDialog (DD) datasets where ASEM outperformed prior models (MoEL/CASE) on many fronts, like 6.2\% in emotion detection. The model produced more coherent responses when evaluated on empathy, fluency, and overall human coherence, Understanding, and empathy on responses given by the model. The results justify that the hierarchical sentiment to emotion analysis model permits a more refined and appropriate understanding and empathic response.

3. Hazarika et al. \cite{hazarika-etal-2018-icon} proposed the Interactive Conversational memory Network (ICON) for multimodal emotion detection in dyadic conversations. They designed a hierarchical memory network which models two distinct emotional influences.As the speaker talks, the emotional inertia is generated through self-influence and the conversation partner also affects the emotion of the speaker through inter-speaker influence. For every utterance, the model first retrieves the features from the three modalities: the language (text transcripts), audio, and video. The model then employs a two-step process to handle the history of the conversation. The first step has a Self-Influence Module (SIM) which contains separate Gated Recurrent Units (GRUs) for each speaker's history, and the second step has a Dynamic Global-Influence Module (DGIM) which has another GRU that forms a global memory bank capturing the interplay of both speakers. The attention mechanism proposed in the model processes a multi-hop access over the conversation memories and crafts a contextual summary that is emotion-classified for the target utterance. The model was tested on the SEMAINE and IEMOCAP datasets (discrete emotion classification). Compared to the legacy models, ICON achieved the highest performance on the IEMOCAP dataset: a weighted average accuracy of 64.0\%, and an F1 score of 63.5, surpassing the other models TFN and MFN. An ablation study has demonstrated the significance of modeling inter-speaker dynamics as well as the necessity of the multi-hop memory mechanism.

These studies reflect the growing research on emotion detection, and the multiple ways and methods used to study and analyze emotions across different modalities.

\section{Methodology}
\subsection{Dataset Description and Preprocessing}

The study uses the DAIR-AI emotion dataset that consists of English tweets annotated with the basic emotions of anger, fear, joy, love, sadness, and surprise and comprises about 20,000 instances. The dataset featured severe class imbalance since the surprise emotion was the least represented. The original distribution of tweets was as follows:
\begin{itemize}
    \item Joy (6,761), Love (1,641), Sadness (5,797), Anger (2,709), Fear (2,373), surprise (701) examples.
\end{itemize}

As the surprise emotion is underrepresented in the dataset, it resembles a unique challenge. To alleviate class imbalance, the class-weighted cross-entropy loss function was utilized, which imposes a higher penalty for misclassification of an example in the surprise class so that the model is motivated to learn the defining characteristics of the class. 
\begin{equation}
    L_{\text{WCE}} = -\frac{1}{N} \sum_{i=1}^{N} \sum_{c=1}^{M} w_c \cdot y_{i,c} \log \left(\hat{y}_{i,c}\right)
\end{equation}

where:
\begin{minipage}[t]{0.48\textwidth}
\begin{itemize}
    \item $L_{\text{WCE}}$ : Weighted Cross-Entropy loss.
    \item $N$ : Total number of samples.
    \item $M$ : Total number of classes.
    \item $i$ : Index of a sample ($i = 1, 2, \dots, N$).
\end{itemize}
\end{minipage}\hfill
\begin{minipage}[t]{0.48\textwidth}
\begin{itemize}
    \item $c$ : Index of a class ($c = 1, 2, \dots, M$).
    \item $y_{i,c}$ : Ground-truth indicator (1 if sample $i$ belongs to class $c$, otherwise 0).
    \item $\hat{y}_{i,c}$ : Predicted probability that sample $i$ belongs to class $c$.
    \item $w_c$ : Weight assigned to class $c$ to account for class imbalance.
\end{itemize}
\end{minipage}

The class weight $w_c$ is computed as:
\begin{equation}
    w_c = \frac{N}{M \cdot N_c}
\end{equation}

where: $N_c$ : Number of samples belonging to class $c$.

For minority class ( the `surprise` class $c=s$ with low $N_s$), the weight $w_s$ is substantially larger compared to frequent classes. This scaling increases the gradient contribution of rare samples during backpropagation, encouraging the model to allocate sufficient capacity for learning underrepresented classes.

\subsection{Emotion Classification}

\begin{figure*}[h]
    \centering
    \includegraphics[width=\textwidth]{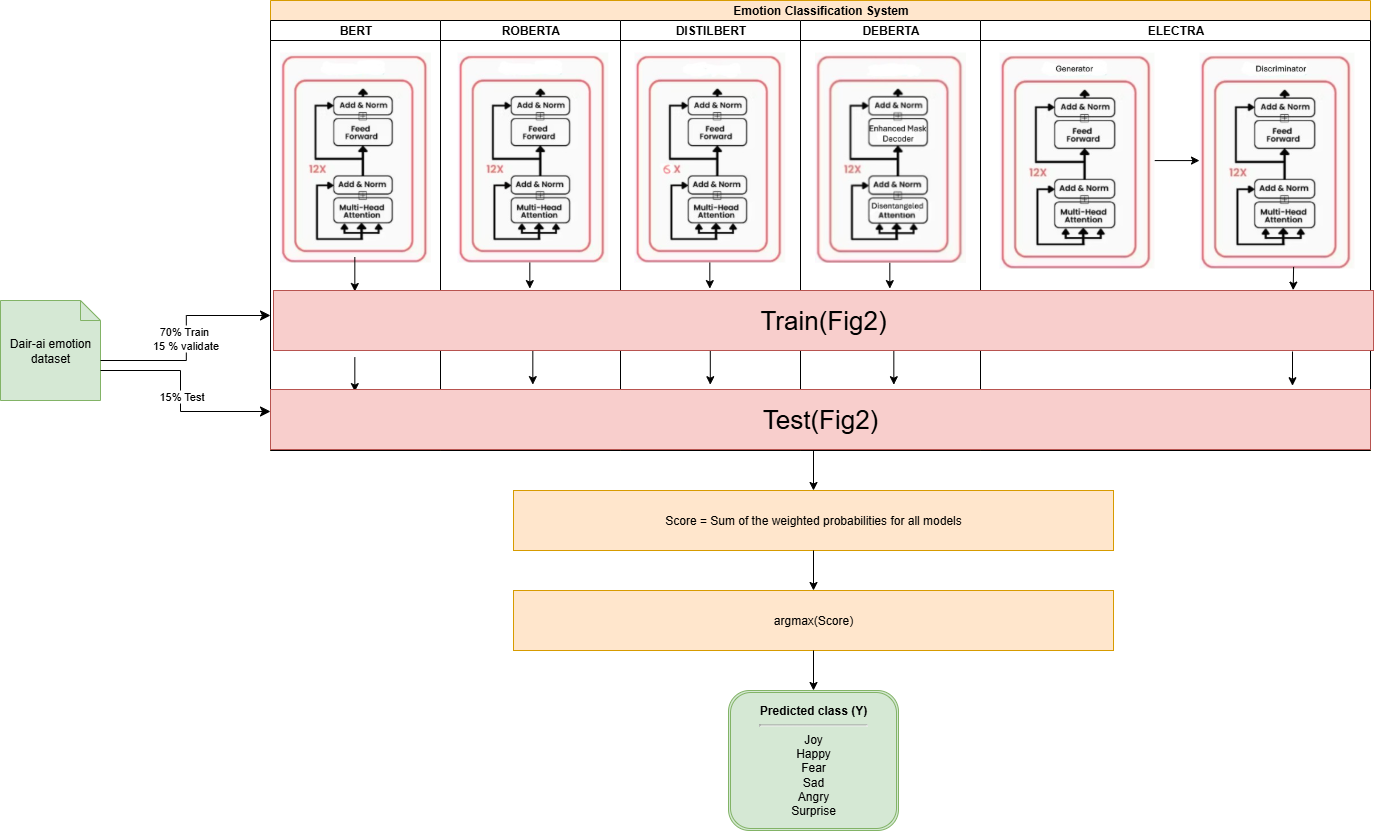}
    \caption{System Architecture.}
    \label{fig:myimage}
\end{figure*}

\begin{figure*}[h]
    \centering
    \includegraphics[width=\textwidth]{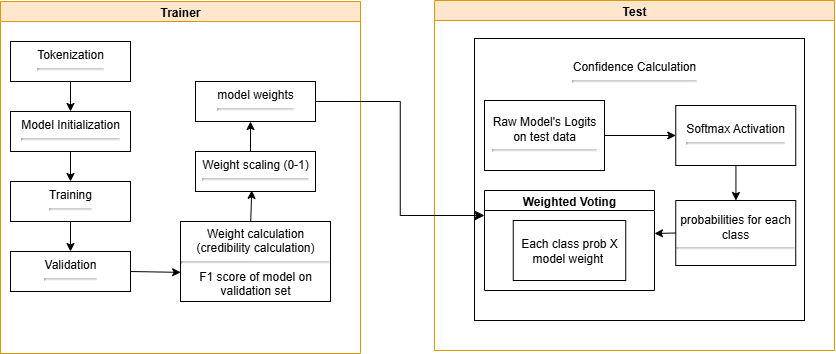}
    \caption{Trainer and test architecture.}
    \label{fig:myimage}
\end{figure*}

The proposed system architecture, illustrated in Figure 1, integrates five fully fine-tuned BERT-family models (BERT, RoBERTa, DistilBERT, DeBERTa, and ELECTRA) through a credibility-confidence weighted voting ensemble to achieve robust emotion classification.

The dataset was split into (70\%), (15\%), and (15\%) training, validation, and test  sets, respectively while maintaining the class distribution within each split. Subsequently, each model was fine-tuned individually on the preprocessed training set. The hyperparameters used in this fine-tuning process as summarized in Table 1, were selected to maximize performance and prevent overfitting.

\begin{table}[H]
\centering
\caption{Hyperparameters used for fine-tuning sLLMs}
\label{tab:hyperparameters}
\begin{tabular}{|l|l|}
\hline
\textbf{Hyperparameter} & \textbf{Value} \\ \hline
Number of Epochs & 4 \\ 
Batch Size (Train/Eval) & 32 \\ 
Learning Rate & $2\times 10^{-5}$ \\ 
Warmup Steps & 500 \\ 
Weight Decay & 0.01 \\ 
Max Sequence Length & 256 \\ 
Mixed Precision (fp16) & Enabled \\ \hline
\end{tabular}
\end{table}

• Number of Epochs (4): The model was trained for 4 epochs. This provided sufficient learning iterations to understand the relevant patterns for emotions without the risk of overfitting, as confirmed by the validation results.

• Batch Size (32): A batch size of 32 was adopted for both the training and evaluation phases, ensuring reasonable computation and that the gradient estimates were stable to foster more reliable convergence.

• Learning Rate (2e-5): The learning rate  $2 \times 10^{-5}$ allowed gradual, stable fine-tuning of pre-trained weights—small enough to prevent catastrophic forgetting yet large enough for efficient task adaptation.

• Warmup Steps (500): A linear warmup over the first 500 steps gradually increased the learning rate to its target value, preventing unstable updates in early training and improving convergence stability.

• Weight Decay (0.01): A weight decay of 0.01 served as L2 regularization, penalizing large weights to enhance generalization and reduce overfitting.

• Maximum Sequence Length (256):  Adaptively, texts were padded or truncated to 256 tokens. This method attended to nearly all of the training data to maintain computational efficiency.
• Mixed Precision Training (fp16): this improved the training speed by using 16-bit arithmetic for most operations.

The adjustment of these hyperparameters has been in accordance with the recommendations from the original model documentation \cite{Morris2020}.

Following training, the validation set was used to compute the F1-score for all trained models. For each model, the F1-score is the estimate of the model’s credibility weight and its generalization ability ($Q_i$), in the ensemble, ($Q_i$) is important as it shows the model’s reliability with respect to the dataset.

During inference, the ensemble performs weighted voting based on the credibility and confidence, synthesizing the predictions of all five models. For the input text x, the score the ensemble produces for the candidate class c is a weighted sum as follows:
\begin{equation}
\text{Score}(c) = \sum_{i=1}^{N} C_i^c(x) \cdot Q_i
\end{equation}

Where: $N$ is the number of models in the ensemble, $C_i^c(x)$ is the predicted probability (confidence) that model $i$ assigns to class $c$ for input $x$, and $Q_i$ is the credibility weight of model $i$, defined as its macro F1-score on the validation set.

The prediction of the ensemble is taken from the class with the highest final score as follows:

\begin{equation}
\hat{y} = \arg\max_c \text{Score}(c)
\end{equation}

This method ensures that models demonstrating higher generalizability on the validation data exert a greater influence on the final decision.

Predictions were made using this credibility-weighted voting scheme and the primary test metric employed was the F1 score as the F1 score is the harmonic mean of Precision and Recall, providing a single metric that balances the two concerns. It is particularly valuable in scenarios with imbalanced class distributions (like our case), as it prevents a model from achieving a high score by simply favoring the majority class. The formulas are as follows:

Precision measures the accuracy of positive predictions: What proportion of predicted positives are truly positive?

\begin{equation}
\text{Precision} = \frac{TP}{TP + FP}
\end{equation}
Recall measures the model's ability to find all positive instances: What proportion of actual positives were correctly identified?
\begin{equation}
\text{Recall} = \frac{TP}{TP + FN}
\end{equation}
F1 Score harmonizes these two metrics:
\begin{equation}
F_1 = 2 \cdot \frac{\text{Precision} \cdot \text{Recall}}{\text{Precision} + \text{Recall}}
\end{equation}
Where TP is True Positives, FP is False Positives, and FN is False Negatives.

\section{Results}
\subsection{Performance of the Proposed Ensemble Model}

The proposed credibility-weighted ensemble model, which integrates BERT, RoBERTa, DistilBERT, DeBERTa, and ELECTRA, registered F1 score of 93.5\% which is state of the art on the emotion classification task. This is significant when compared to the performance of several large language models (LLMs). To have a comprehensive and fair comparison, 2 evaluation paradigms, as illustrated in Table 2, were set for the LLMs:
\begin{itemize}
    \item Evaluation without any finetuning (a zero-shot approach)
    \item Evaluation in which the model was trained on the task using Low-Rank Adaptation (LoRA)
    
\end{itemize}

\begin{table}[H]
\centering
\caption{Performance Comparison: Proposed Ensemble vs. LLMs (Zero-Shot and LoRA-Adapted)}
\label{tab:results}
\begin{tabular}{|l|c|c|c|}
\hline
\textbf{Model} & \textbf{Zero-Shot F1 (\%)} & \textbf{LoRA-Adapted F1 (\%)} & \textbf{Total Parameters} \\
\hline
\textbf{Ours} & \textbf{-} & \textbf{93.50} & \textbf{$\sim$595M} \\
\noalign{\vskip 0.5ex} 
\hline
Qwen-1.8B        & 52 & 93.20 & 1.8B \\
Falcon-7B        & 63 & 91.47 & 7.0B \\
Mistral-7B       & 66 & 88.02 & 7.1B \\
Phi-2 (2.7B)     & 50 & 91.30 & 2.7B \\
OpenLLaMA-3B     & 48 & 86.13 & 3.0B \\
\hline
\end{tabular}
\end{table}

As highlighted in Table 2, the ensemble not only outperforms all zero-shot LLMs by a significant margin, but also surpasses the same LLMs after task-specific training using a consistent LoRA setup on the task (rank=8, lora alpha=16, lora dropout=0.1). This is a remarkable achievement considering outperforming the best LoRA-adapted LLM (Qwen at 93.2\%) while being vastly more parameter efficient.

\subsection{Analysis of Parameter Efficiency}

A critical finding of this study is the exceptional parameter efficiency demonstrated by the proposed ensemble. The combined parameter count of all five constituent models (BERT-110M, RoBERTa-125M, DistilBERT-66M, DeBERTa-100M, ELECTRA-33M) is approximately 595 million (0.595B). This is substantially smaller than even the most parameter-efficient LLM in the comparison, Qwen (1.8B), which has about three times the number of parameters.

The performance superiority of the proposed model becomes even more pronounced when considering the scale of other competitors. The ensemble, with under half a billion parameters, not only surpasses the 1.8-billion-parameter model like Qwen but also edges out the 7 billion-parameter models like Mistral and Falcon. This shows that an ensemble of smaller, fully fine-tuned transformer models, even at the smaller end, can be tailored to achieve better task specific performance compared to much larger, general-purpose LLMs. The larger general purpose LLMs tend to perform worse on specialized tasks because of the recitation curse and lack of prompt engineering sophistication, which is required to get a similar performance on a specialized task.

This result underscores a key insight: for specialized NLP tasks such as emotion classification, exhaustive fine-tuning of well structured, smaller models on a dataset can be a more effective and computationally efficient strategy than relying on the raw, large scaled capabilities of LLMs. The success of the ensemble is attributed to the complementary shared expertise of its members and the credibility-weighted voting system that best harnesses their proven task-specific capabilities.

\subsection{ Benchmark Comparison with State-of-the-Art}
To further illustrate the efficacy of our proposed method, we compare our results to a recent state-of-the-art study focused on the DAIR-AI emotion dataset, and published in a Q1 journal \cite{BARCENARUIZ2025114070}. This is particularly insightful as both pieces of work have similar methodological approaches and each implements a new weighted voting mechanism while both operate within the CJT framework.

According to the benchmark study results, the maximum macro F1-score was 93.7\%. Our credibility-weighted voting ensemble, as shown in the Table 3, achieves the macro F1-score of 93.50\% on the same dataset and task, which is still very competitive.

The remarkable proximity of these results—with a minimal difference of only 0.2\% demonstrates that our method is highly comparable to the state-of-the-art methods available. Our results closely approach these methods, which confirms the performance of the credibility-weighted voting method toward emotion classification tasks, and further supports the use of it as an alternative ensemble method for emotion classification. The results obtained further prove the effectiveness of ensemble methods for this natural language processing problem.

\begin{table}[h!]
\centering
\caption{Performance Comparison on DAIR-AI Dataset with Q1 Journal}
\renewcommand{\arraystretch}{1.2} % increase row height
\begin{tabular}{lc}
\hline
\textbf{Model/Approach} & \textbf{F1 Score} \\
\hline
BRTO1 & 89.5 \\
BRTO2 & 90.1 \\
ROBO1 & 92.6 \\
ROBO2 & 92.9 \\
DISO1 & 89.6 \\
Average (Benchmark) & 90.9 \\
JC & 93.7 \\
JA & 93.7 \\
JD & 93.7 \\
Ours & 93.5 \\
\hline
\end{tabular}
\end{table}

Another comparison was conducted against a paper published in ACL workshops\cite{kermani2025systematic} which investigated various paradigms for emotion classification on the same DAIR-AI Emotion dataset.

Table 4 shows the performance results of the model from the referenced study \cite{kermani2025systematic} in which the highest result was (87\%), thst was achieved using a fine-tuning approach. Their other approaches, including Zero-shot, Few-shot, and Retrieval-Augmented Generation (RAG), resulted in much lower results.

In contrast, in this work we demonstrate that the credibility-weighted ensemble model achieves a macro F1-score of 93.5\%, which is 6.5\% higher than the previous fine-tuning score on the same dataset and benchmark. This score affirms the value of the ensemble approach and sets a new standard in the DAIR-AI Emotion classification task in that benchmark.

\begin{table}[h]
\centering
\caption{Performance comparison with ACL Workshop Paper}
\label{tab:acl_comparison}
\begin{tabular}{|l|c|c|}
\hline
\textbf{Method} & \textbf{F1-Score} \\
\hline
\textbf{Ours}  & \textbf{0.935} \\
\hline
Fine-tuning  & 0.87 \\
Zero-shot    & 0.38 \\
Few-shot     & 0.30 \\
RAG          & 0.32 \\
\hline
\end{tabular}
\end{table}

\section{Limitations and Future Work}
While this study does offer a comprehensive framework for emotion classification and presents a competitive score, there are limitations within this work, which of course, provide avenues for future work.

1.Dataset and Linguistic Specificity: The analyses were performed on the DAIR-AI dataset of English tweets only. The performance and conclusions may not generalize to other languages, domains, or platforms with different linguistic characteristics.

2.Focus on a Homogeneous Model Family:
This research concentrated exclusively on models from the BERT-family architecture. While this provided a controlled and highly effective foundation, it inherently limited the exploration of diversity that could be introduced by incorporating top-performing models from other architectural lineages. Future work should expand the ensemble paradigm to be architecture-agnostic. It could investigate the integration of models from diverse architectural families into the credibility-weighted voting framework. This would test the hypothesis that increasing architectural diversity, in addition to data diversity, could lead to even more robust and generalized ensembles.

\section{Conclusion}
This study has presented a comprehensive investigation into ensemble strategies for emotion classification. Our work makes a key contributions to the field as, we established the superiority of emotion classification using BERT-family models with full fine-tuning, achieving competitive results on the DAIR-AI dataset. Most significantly, our proposed credibility-weighted voting ensemble consistently outperformed all LLMs models, achieving a 93.50\% F1-score. This proves that our ensemble method successfully integrates various strengths of the models while reducing, to some extent, their weaknesses.

The key finding of this research is that this performance was achieved not by increasing the size of the model but through the innovative use of parameter-efficient frameworks. Our ensemble, which totalled around 595 million parameters, was able to surpass the performance of numerous other larger Large Language Models (LLMs) including 7B parameter models such as Falcon and Mistral, and even after those models were task-adapted using LoRA fine-tuning. This finds that sophisticated ensembles of smaller, specialized models, rather than larger, generalized LLMs, can better serve the intended purpose of NLP tasks, therefore contesting the theory that performance correlates directly and solely with the number of parameters.

Our findings have critical practical consequences. For researchers and practitioners, our results can serve as a blueprint for designing smaller models to achieve high performance, thus avoiding the high computational demands associated with deploying LLMs. The proposed confidence-credibility weighting mechanism offers a principled, interpretable, and flexible approach for aggregating model predictions, which can be directly applied to a wide range of text classification problems beyond emotion recognition.

\label{sect:bib}
\bibliographystyle{plain}
\bibliography{easychair}

%------------------------------------------------------------------------------
\end{document}